\documentclass[twoside]{article}

\usepackage{amsmath,amssymb}
\usepackage{tikz}
\usepackage{booktabs}
\usepackage{caption}
\usepackage{subcaption}
\usepackage{graphicx}
\usepackage{algorithm}
\usepackage{algpseudocode}
\usepackage{listings}
\usepackage{algorithm}
\usepackage{algpseudocode}
\usepackage{xcolor}
\usepackage[T1]{fontenc}
\usepackage[utf8]{inputenc}
\usepackage{hyperref}
\usepackage[accepted]{aistats2020}
\usepackage[round]{natbib}

\begin{document}

%

%

\newcommand{\quotes}[1]{``#1''}

\twocolumn[
    \aistatstitle{A Deep Generative Model for Fragment-Based Molecule Generation}
    \aistatsauthor{Marco Podda \And Davide Bacciu \And  Alessio Micheli }
    \aistatsaddress{University of Pisa,\\Largo Bruno Pontecorvo 3,\\56127 Pisa, Italy \And University of Pisa,\\Largo Bruno Pontecorvo 3,\\56127 Pisa, Italy \And University of Pisa,\\Largo Bruno Pontecorvo 3,\\56127 Pisa, Italy} 
]

\begin{abstract}
Molecule generation is a challenging open problem in cheminformatics. Currently, deep generative approaches addressing the challenge belong to two broad categories, differing in how molecules are represented. One approach encodes molecular graphs as strings of text, and learns their corresponding character-based language model. Another, more expressive, approach operates directly on the molecular graph. In this work, we address two limitations of the former: generation of invalid and duplicate molecules. To improve validity rates, we develop a language model for small molecular substructures called fragments, loosely inspired by the well-known paradigm of Fragment-Based Drug Design. In other words, we generate molecules fragment by fragment, instead of atom by atom. To improve uniqueness rates, we present a frequency-based masking strategy that helps generate molecules with infrequent fragments. We show experimentally that our model largely outperforms other language model-based competitors, reaching state-of-the-art performances typical of graph-based approaches. Moreover, generated molecules display  molecular properties similar to those in the training sample, even in absence of explicit task-specific supervision.
\end{abstract}

\section{\uppercase{Introduction}}\label{sec:introduction}
The term \emph{de novo} Drug Design (DD) refers to a collection of techniques for the production of novel chemical compounds, either by \emph{in-vitro} synthesis or computer-aided, endowed with desired pharmaceutical properties. Among synthesis-based methodologies of DD, Fragment-Based Drug Design (FBDD) \citep{fbdd, fbdd2} has established itself as an effective alternative to more traditional methods such as High-Throughput Screening (HTS). At the core of FBDD is the notion of fragments, small molecular weight compounds that are easily synthesizable, have high binding affinity and weakly interact with a set of target molecules. Fragments are combined together according to several strategies, producing more complex compounds with enhanced target interactions.

In contrast to synthesis-based methods, computational approaches to DD are based on the efficient exploration of the space of molecules, which is an inherently hard problem because of its size (estimated to be in the order of $10^{60}$). Recently, deep generative models of molecules have shown promising results in this challenging task \citep{xu-molecular-generation-review}. 

Broadly speaking, deep learning models for molecule generation are typically based on an encoder-decoder approach. First, the molecular graph is encoded in a vectorial latent space.  Then, a decoding distribution is placed on such latent codes, which is subsequently exploited for efficient sampling. Depending on which input representation of the molecular graph is chosen, we distinguish two broad families of approaches. The first family of models uses a textual representation of the molecular graph, e.g. the SMILES \citep{smiles} language, where atoms and chemical bonds are represented as characters. From this representation, a character-based language model (LM) \citep{sutskever-char-based-language-models} can be trained using Recurrent Neural Network (RNN) \citep{elman-rnn} architectures. For this reason, we term approaches of this kind as \emph{LM-based}. The second family operates directly on the molecular graph, encoding it either sequentially using RNNs, or in a permutation-invariant fashion using Graph Neural Networks (GNN) \citep{gnn-scarselli,gnn-micheli}. We term this family of models \emph{graph-based}. Both approaches have advantages and disadvantages: for example, graph-based models are more expressive in principle, because they act directly on the molecular graph. However, they are hard to train and less efficient to sample from. In contrast, LM-based approaches trade-off a less expressive intermediate representation of the molecular graph with efficient training and sampling. Another common issue with LM-based approaches is that they tend to generate a large share of chemically invalid molecules, as well as many duplicates of the most likely molecules. For these reasons, graph-based methods typically hold state-of-the-art performances as regards the production of chemically valid, novel and unique molecules.

In this work, we address the two main shortcomings of LM-based generative models. Our first contribution is to counter low validity rates. To this end, we take inspiration from FBDD and develop a fragment-based language model of molecules. In other words, instead of generating a molecule atom by atom, we generate it \emph{fragment by fragment}. Note that since fragments are chemically sound, our approach needs to ensure validity only when connecting a novel fragment; in contrast, character-based LM approaches need to maintain validity after each novel atom is added. Hence, our approach naturally ensures higher validity rates.

As a second contribution, we develop a simple strategy that fosters the generation of unique molecules, avoiding duplicates. In our fragment-based framework, the problem of duplicates is a consequence of the distribution of fragments in the data. Roughly speaking, the distribution of fragments follows a power-law distribution, with a small number of very frequent fragments as opposed to a large number of infrequent fragments. Thus, we mask infrequent fragments with a token that specifies their frequency. During generation, whenever the masking token is predicted, we sample from the set of fragments that were masked by that token.

Our experimental evaluation shows that our model is able to perform on par with state-of-the-art graph-based methods, despite using an inherently least expressive representation of the molecular graph. Moreover, we show that generated compounds display similar structural and chemical features to those in the training sample, even without the support of explicit supervision on such properties.

\section{\uppercase{A primer on fragments}}
Here, we briefly describe what are fragments and how they are used in the context of DD. Fragments are very-small-weight compounds, typically composed of $<20$ non-hydrogen atoms. Small size has several advantages: firstly, they are easier to manipulate chemically than larger fragments. Secondly, the chemical space of fragments is narrower than, for example, the one of drug-like molecules typically generated from other DD approaches such as HTS. Thus, it is easier to explore and characterize. Thirdly, the small size makes fragments weakly interact with a broader spectrum of target proteins than larger compounds (higher molecular complexity translates into strongest interaction, albeit not necessarily beneficial). 

A typical FBDD experiment begins with the identification of a suitable collection of fragments, from which a subset with desired interactions with the target (hits) is identified. Subsequently, fragments are optimized into higher affinity compounds that become the starting points (leads) for subsequent drug discovery phases. Optimization is commonly carried out according to three different strategies: \emph{a}) linking, which optimizes a given fragment by connecting it with another fragment; \emph{b}) growing, where the fragment is functionally and structurally enriched to optimize binding site occupation; \emph{c}) merging, which involves combining the structure of two overlapping fragments into a new one with increased affinity.
Since its inception in 1996, FBDD accounts for two clinically approved drugs, and more than thirty undergoing clinical trials at various stages \citep{fbdd3}.

\begin{figure*}
\centering
\begin{minipage}{0.4\textwidth}
\centering
\includegraphics[height=223px]{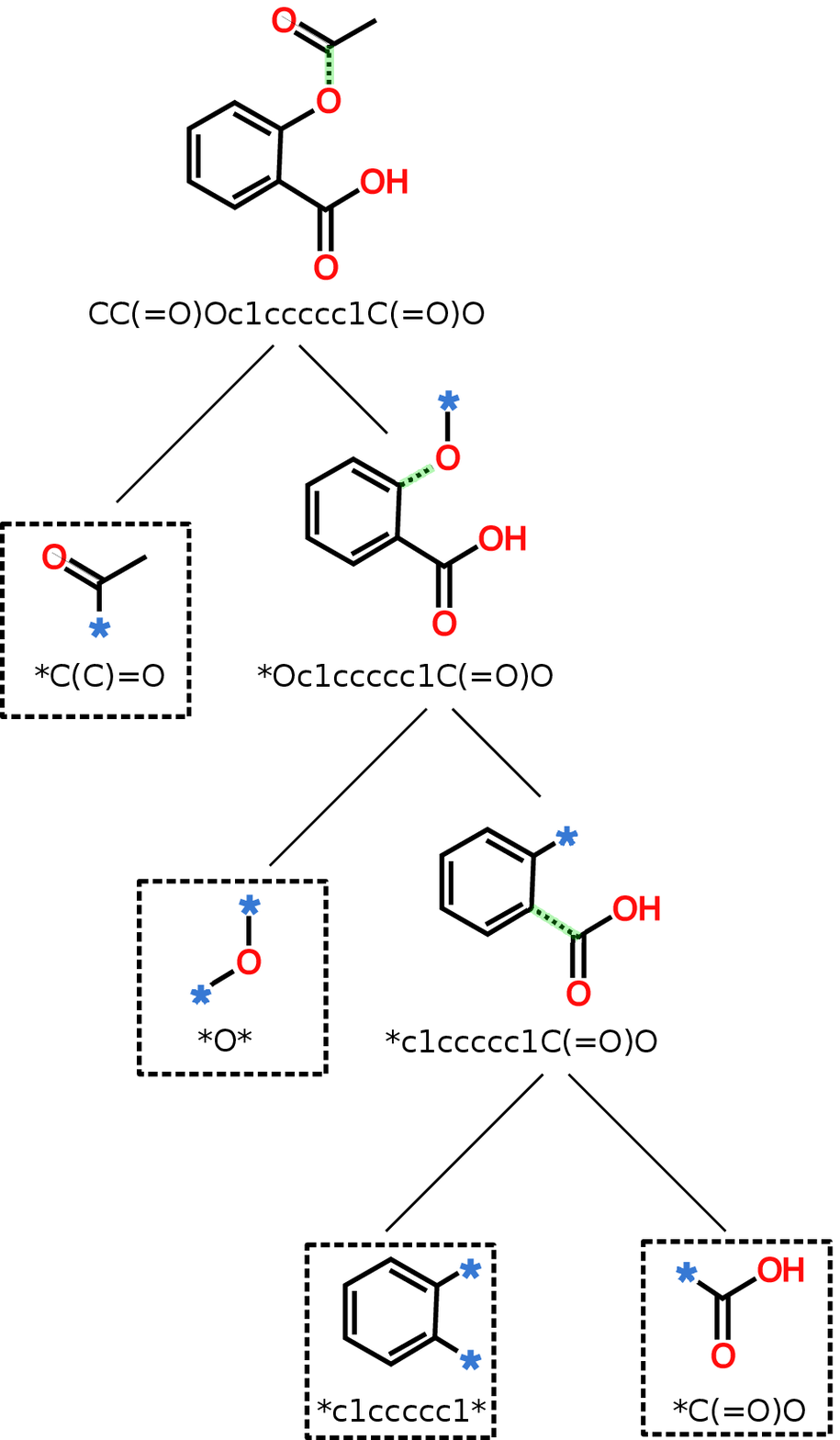}
\end{minipage}
\begin{minipage}{0.49\textwidth}
\centering
\begin{algorithm}[H]
        \caption{Fragmentation}
        \begin{algorithmic}[1]
        
        \Require Molecule $M$, Fragment List $F \leftarrow[]$
        \Procedure{Fragment}{$M, F$}       
            \State \textbf{declare} Bond $b$
            \State $b \leftarrow$ \textsc{GetFirstBRICSBond}$(M)$
            \If{\textsc{IsEmpty($b$)}} 
                \State \Return $F$ 
            \Else
                \State \textbf{declare} Fragment f
                \State \textbf{declare} Molecule $M'$
                \State $f$, $M' \leftarrow$ \textsc{BreakMolAtBond}($M, b$)
                \State $F \leftarrow$ \textsc{Append}($F$, $f$)
                \State $M \leftarrow M'$
                \State \textsc{Fragment}($M, F$)
            \EndIf
        \EndProcedure
        \end{algorithmic}
        \label{algo:fragmentation}
        \end{algorithm}
\end{minipage}
    \caption{Left: a depiction of the fragmentation procedure. The root of the tree is the molecule to be fragmented (aspirin), while the leaves (enclosed by dashed boxes) represent the extracted fragments. At each iteration (level), the molecule atoms are scanned from left to right according to the SMILES ordering, extracting a fragment as soon as a breakable bond is found. The process is repeated until the remaining fragment cannot be split further. To reconstruct a molecule, fragments are reassembled starting from the leaves to the root, right to left. Asterisks denote dummy atoms. The dashed bonds with the green highlight are the ones selected to be broken/joined using BRICS rules. Right: a sketch of recursive implementation of the fragmentation algorithm.}
    \label{fig:fragmentation}
\end{figure*}

\section{\uppercase{Related Works}}
In contrast with general-purpose models which use auto-regressive generation to sample novel graphs \citep{you-graphrnn,podda-graph-generation}, molecular generators are usually arranged in an encoder-decoder scheme, coupled with a generative model that is trained to learn the distribution of codes in latent space, either explicitly using variants of Variational Auto-Encoders (VAEs) \citep{vae} or implicitly using Generative Adversarial Networks (GANs) \citep{gan}. Novel molecules can be generated by sampling the latent space, and letting the decoder reconstruct the molecular graph, conditioned on the sampled code. We now adopt the taxonomy of Section~\ref{sec:introduction} and recap approaches belonging to the LM-based as well as graph-based families. We especially focus on VAE-based models, as of direct relevance for this work. 

\paragraph{Language Model-Based Approaches} A seminal LM-based model for molecular generation is the work of \cite{chemvae}, which is essentially a character-based language model of SMILES strings coupled with a VAE to learn the distribution of the latent space. A first extension to constrain the generation with syntactic rules is proposed in the work of \cite{grammarvae}. 
The work of \cite{sdvae} extends this approach further, augmenting the VAE generator with a form of syntax-directed translation, thus ensuring that generated molecules are both syntactically valid, as well as semantically reasonable. Notice that our work is LM-based, but differs from existing approaches in that we do not generate a molecule atom by atom, but rather fragment by fragment.
\paragraph{Graph-Based Approaches} Early contributions in this line of research were based on encoding the molecular graph with various strategies, and decoding its adjacency matrix directly. For example, \cite{graphvae} use a VAE-based architecture with a GNN encoder. The decoder outputs a probabilistic fully-connected graph, where the presence of an edge is modeled as a Bernoulli process, assuming edge independence. The final graph is sparsified with approximate graph matching. 
\cite{nevae} propose a different approach. Similarly to the work of \cite{graphvae}, a GNN is used as encoder; however, only node embeddings are mapped to latent space. The decoder works by first sampling a set of atom embeddings and inferring their type from a categorical distribution. Then, bonds between all possible pairings of such atoms are predicted, and their their type is inferred from another categorical distribution. The whole architecture is end-to-end trainable. 
A similar approach is developed in the work of \cite{cgvae}, where a Gated-Graph Neural Network \citep{ggnn} is used as encoder. The decoder first samples a set of nodes, then sequentially adds the edges for each node on the basis of a breadth-first queue. 
Finally, the model by \cite{jtvae} generates molecules by first sampling a tree structure that specifies how functional pieces of the molecule are connected. Then, it uses the sampled tree to predict the molecular subgraphs corresponding to each tree node.
\begin{figure*}
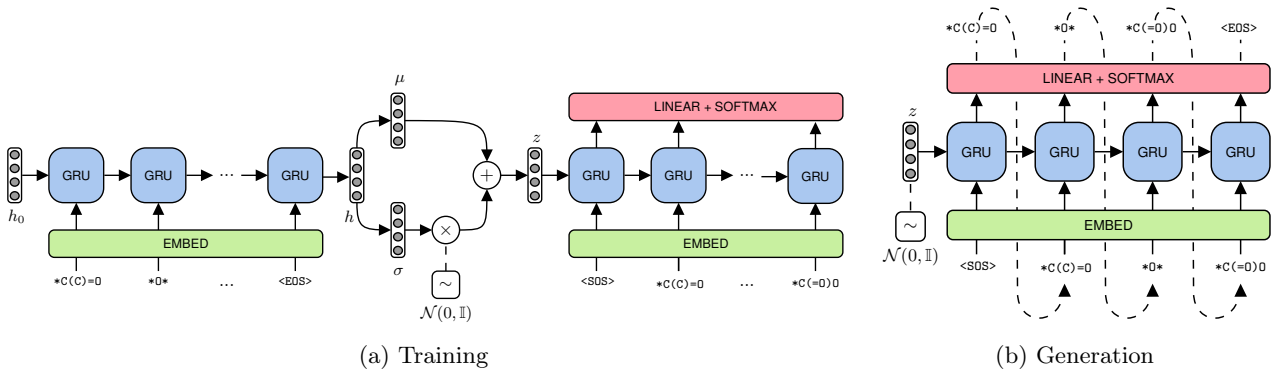

     \centering
     \begin{subfigure}[b]{0.67\textwidth}
         \centering
         \resizebox{0.99\textwidth}{!}{\input{figs/training.tex}}
         \caption{Training}
         \label{fig:model-training}
     \end{subfigure}
     \hfill
     \begin{subfigure}[b]{0.32\textwidth}
         \centering
         \resizebox{0.99\textwidth}{!}{\input{figs/generation.tex}}
         \caption{Generation}
         \label{fig:model-generation}
     \end{subfigure}
    \caption{The proposed architecture during training (a) and generation (b). The EMBED layer (in green) is the skip-gram embedding matrix of the textual representation of fragments; the GRU layers (in blue) are the recurrent units that encode and decode fragments; the LINEAR + SOFTMAX (in red) layers serve the purpose of projecting the GRU outputs to the space of the vocabulary, and computing the probability of the next fragment, respectively. Dashed lines indicate sampling.}
    \label{fig:model}
\end{figure*}
\section{\uppercase{Methods}}
At a high level, our approach encompasses three steps: break molecules into sequences of fragments, encode them as SMILES words, and learn their corresponding language model. In this section, we review each steps and provide the necessary details on how we operated.
\subsection{Molecule Fragmentation}
Given a dataset of molecules, the first step of our approach entails breaking them into an ordered sequence of fragments. To do so, we leverage the Breaking of Retrosynthetically Interesting Chemical Substructures (BRICS) algorithm \citep{brics}, which breaks strategic bonds in a molecule that match a set of chemical reactions. "Dummy" atoms (with atomic number 0) are attached to each end of the cleavage sites, marking the position where two fragments can be joined together. BRICS cleavage rules are designed to retain molecular components with valuable structural and functional content, e.g. aromatic rings and side-chains, breaking only single bonds that connect among them. Our fragmentation algorithm works by scanning atoms in the order imposed by the SMILES encoding. As soon as a breakable bond (according to the BRICS rules) is encountered during the scan, the molecule is broken in two at that bond, applying a matching chemical reaction. After the cleavage, we collect the leftmost fragment, and repeat the process on the rightmost fragment in a recursive fashion. Note that fragment extraction is ordered from left to right according to the SMILES representation; this makes the process fully reversible, i.e. it is possible to reconstruct the original molecule from a sequence of fragments. In Figure~\ref{fig:fragmentation}, we show a practical example of the fragmentation process and provide a pseudo-code recursive implementation of our algorithm.
\subsection{Fragment Embedding}
The former process transforms a dataset of molecules into a dataset of sequences of SMILES-encoded fragments. In analogy with the work of \cite{bowman-sentences-continuous-space}, we view a sequence of fragments as a \quotes{sentence}; therefore, we construct a vocabulary of unique fragment \quotes{words}. We embed each fragment by pushing fragments that occur in similar contexts to be mapped to similar regions in embedding space. More formally, given a sequence $s = ( s_1, s_2, \ldots s_{|s|})$  of SMILES-encoded fragments, we minimize the following objective function:
$$\mathcal{L}(s) = - \sum_{i=1}^{|s|}\sum_{-w \leq j \leq w} \log P(s_{i+j}| s_i),\ j \neq 0,$$
where $w$ is the size of the context window, $s_i$ is the target fragment, and $s_{i+j}$ are context fragments. In this work, $P$ is implemented as a skip-gram model with negative sampling \citep{mikolov-skipgram}. After training the embeddings, each fragment sequence is represented as $x = (x_1, x_2, \ldots, x_{|x|})$, where the generic $x_i$ is a column vector of the skip-gram embedding matrix.
\subsection{Training}
Similarly to other language models, we adopt an encoder-decoder architecture with a generative model in between the two. Here, we describe the architecture and the training process in detail.
\paragraph{Encoder} To encode the sequence of fragments, we use Gated Recurrent Units (GRUs) \citep{gru}. Specifically, we transform each embedding $x_i$ into a hidden representation $h_i = \mathrm{GRU}(x_i, h_{i-1})$ as follows\footnote{We omit bias terms for clarity.}:
\begin{equation}\label{eq:gru}
\begin{split}
r_i &= \mathrm{sigmoid}(W_r x_i + U_r h_{i-1})\\
u_i &= \mathrm{sigmoid}(W_u x_i + U_u h_{i-1})\\
v_i &= \mathrm{tanh}(W_h x_i + U_h(r_i \odot h_{i-1}))\\
h_i &= u_i \odot h_{i-1} + (1-u_i) \odot v_i,
\end{split}
\end{equation}
where $h_0$ is the zero vector. In the above formula, $r_i$ is a reset gate vector, $u_i$ is an update gate vector; $W$ and $U$ are weight matrices, and $\odot$ denotes element-wise multiplication ($v_i$ is a convenience notation for ease of read). The hidden representation of the last fragment in the sequence, which we term $h$, is used as latent representation of the entire sequence. The encoder is trained to minimize the following Kullback-Leibler (KL) divergence:
$$\mathcal{L}_{\mathrm{enc}}(x) = -\mathrm{KL}(\mathcal{N}(\mu, \mathrm{diag}(\sigma^2))\; ||\; \mathcal{N}(0, \mathbb{I})).$$
In this work, $\mu = W_{\mu}h + b_{\mu}$ and $\log(\sigma^2) = W_{\sigma}h + b_{\sigma}$, where $W$ denotes weight matrices, and $b$ denotes bias terms.

\paragraph{Decoder} The decoder is a recurrent model with GRU units. Its hidden state is initialized by applying the reparameterization trick \citep{vae}, setting $z = h_0 = \mu + \sigma \epsilon$, with $\epsilon \sim \mathcal{N}(0, \mathbb{I})$, as the initial hidden state of the decoder. Differently from the encoder, the decoder also computes the output probability associated to the next element in the sequence as follows:
$$P(x_{i+1}|x_i,h_{i-1}) = \mathrm{softmax}(W_{out} h_i + b_{out}),$$
where $h_i = \mathrm{GRU}(x_i, h_{i-1})$ similarly to the encoder, weight matrix $W_{out}$ projects the hidden representation to the space of the vocabulary, and $b_{out}$ is a bias term. During training, we use teacher forcing \citep{williams-teacher-forcing} and feed the ground truth fragment as input for the following step. The decoder is trained to minimize the negative log-likelihood of the fragment sequence:
$$\mathcal{L}_{\mathrm{dec}}(x) = - \sum_{i=1}^{|x|} \log P(x_{i+1} \mid x_i, h_{i-1}).$$ Note that this loss corresponds to the Cross-Entropy between the one-hot encoded ground truth sequence and the predicted fragment probabilities for each of its elements, computed as described above. 
\paragraph{Model Loss} Our language model is trained in an end-to-end fashion on a dataset of fragment sequences $\mathcal{D}$. The overall loss is the sum of the encoder and decoder losses for each fragment sequence:
$$\mathcal{L}(\mathcal{D}) = \sum_{x \in \mathcal{D}} \mathcal{L}_{\mathrm{enc}}(x) + \mathcal{L}_{\mathrm{dec}}(x).$$
In analogy with the VAE framework, the decoder loss can be viewed as the reconstruction error of the input sequence, while the encoder loss acts as a regularizer that forces the encoding distribution to be Gaussian. Fig.~\ref{fig:model-training} provides an overview of the architecture.
\subsection{Generation} The generative process starts by sampling a latent vector $z \sim \mathcal{N}(0, \mathbb{I})$, which is used as the initial state of the decoder. The first input of the decoder is an \texttt{SOS} token. Tokens and recurrent states are passed through the GRU, linear and softmax layers to produce an output probability for the next fragment. From it, we use a greedy strategy and sample the most likely fragment, which becomes the input of the next decoding step. The generative process is interrupted whenever an \texttt{EOS} token is sampled. The resulting fragment sequence is finally reassembled into a molecule. Note that, for a sequence to be decodable, it is necessary that the first and last fragments contain exactly one attachment point (because they connect only to one fragment), whereas intermediate fragments need to have two (because they are connected to the preceding and following fragments). Sequences which do not respect this constraint are rejected. Fig.~\ref{fig:model-generation} illustrates the generative process.

\subsection{Low-Frequency Masking}
To foster molecule diversity, we start from the observation that the distribution of fragments in the data can be roughly approximated by a power law distribution. In fact, there is usually a small number of fragments with very high frequency, as opposed to a very large number of fragments that occur rarely. Hence, infrequent fragments are unlikely to be sampled during generation. To counter this, we develop a strategy which we term Low-Frequency Masking (LFM). During training, we mask fragments with frequency below a certain threshold $k$ with a token composed of its frequency and the number of attachment points. As an example, suppose that fragment \texttt{*Nc1ccc(O*)cc1} occurs 5 times in the dataset, and the threshold is $k=10$. Thus, this fragment is masked with the token \texttt{5\_2}, where 5 denotes its frequency, and 2 denotes the number of attachment points. Similarly, fragment \texttt{*C(=O)N1CCN(Cc2ccccc2)CC1} with frequency of 3 is masked with the token \texttt{3\_1}. In contrast, fragment \texttt{*c1ccccc1OC} with a frequency of 200 is left unmasked, since its frequency is above the threshold. A reverse mapping from the masking tokens to the masked fragments is kept. During sampling, whenever a masking token is sampled, we replace it with a fragment sampled with uniform probability from the corresponding set of masked fragments. This strategy serves a double purpose. Firstly, it greatly reduces vocabulary size during training, speeding up the computations. Secondly, it fosters molecule diversity by indirectly boosting the probability of infrequent fragments, and injecting more randomness in the sampling process at the same time. From another point of view, LFM forces the model to generate molecules mostly composed of very frequent fragments, but with infrequent substructures that may vary uniformly from molecule to molecule.

\begin{table}[t]
\begin{center}
\caption{Dataset statistics.}\label{tab:statistics}
\scriptsize
\begin{tabular}{lcc}
    \toprule
    
    \textbf{} & \textbf{ZINC} & \textbf{PCBA}\\
    \midrule
    Total number of molecules     & 249455    & 437929\\
    Molecules with no. fragments $\geq 2$  & 227946 & 383790\\
    Mean number of fragments & 2.24$\pm$0.45 & 2.25$\pm$0.48\\
    Vocabulary size & 168537 & 199835\\
    Vocabulary size (LFM) & 21085 & 35949\\
    Average number of atoms & 23.52$\pm$4.29 & 26.78$\pm$6.76\\
    Average number of bonds & 25.31$\pm$5.07 & 28.98$\pm$7.44\\
    Average number of rings & 2.75$\pm$1.00 & 3.16$\pm$1.05\\
    \bottomrule
\end{tabular}
\end{center}
\end{table}

\section{\uppercase{Experiments}}\label{sec:experiments}
Following, we review our experimental setup, namely how experiments are conceived, which dataset and evaluation metrics were used, which baselines we compare to, as well as details about the hyper-parameters of our model. In our experiments, we try to provide an empirical answer to the following questions: 
\begin{itemize}
    \item Q1: is our fragment-based language model able to increase validity rates? 
    \item Q2: is our LFM strategy beneficial to increase uniqueness rates?
\end{itemize}
To answer the first question, we compare our model against character-based baselines, which generate molecules atom by atom. As regards the second question, we perform an ablation study of performances with and without LFM. We also compare against graph-based approaches, to assess performances in relation to models that use more expressive molecule representations.

\subsection{Data}
We experiment on the ZINC dataset \citep{zinc}, consisting of $\approx$ 250k drug-like compounds. ZINC is a common benchmark for the generative task; as such, it is used to compare against several baselines. To assess the impact of LFM further, in our ablation study we also test our model variants with the PubChem BioAssay (PCBA) dataset  \citep{pubchem}, which comprises $\approx$ 440k small molecules. Dataset statistics are presented in Table~\ref{tab:statistics}.
\paragraph{Preprocessing} We applied some common preprocessing steps before training. In the PCBA dataset, we found 10822 duplicate or invalid molecules, which were removed. After fragmentation, we discarded molecules composed of $< 2$ fragments. After preprocessing, our training samples were $227946$ (ZINC) and $383790$ (PCBA). For completeness, we report that we tried to test our model on the QM9 dataset \citep{qm9} as well, but found out that approximately 70\% of its molecules are composed of a single fragment, making assessment poorly informative due to the small sample size. 
\begin{table*}[h]
\begin{center}
\caption{Scores obtained by our model against LM-based and graph-based baselines. LFM indicates that the model has been trained with Low-Frequency Masking. Performances of our LFM variant are shown in bold.}\label{tab:results}
\footnotesize
\begin{tabular}{lccccc}
    \toprule
    \textbf{Model} & \textbf{Model Family} & \textbf{Dataset} & \textbf{Valid} & \textbf{Novel} & \textbf{Unique}\\
    \midrule
    ChemVAE     & LM    & ZINC & 0.170 & 0.980 & 0.310\\
    GrammarVAE  & LM    & ZINC & 0.310 & 1.000 & 0.108\\
    SDVAE       & LM    & ZINC & 0.435 &     - &     -\\
    GraphVAE    & Graph & ZINC & 0.140 & 1.000 & 0.316\\
    CGVAE       & Graph & ZINC & 1.000 & 1.000 & 0.998\\
    NeVAE       & Graph & ZINC & 1.000 & 0.999 & 1.000\\
    \midrule
    Ours        & LM    & ZINC & 1.000 & 0.992 & 0.460\\
    \textbf{Ours (LFM)}  & LM    & ZINC & \textbf{1.000} & \textbf{0.995} & \textbf{0.998}\\
    \midrule
    Ours        & LM    & PCBA & 1.000 & 0.981 & 0.108\\
    \textbf{Ours  (LFM)} & LM  & PCBA & \textbf{1.000} & \textbf{0.991} & \textbf{0.972}\\
    \bottomrule
\end{tabular}
\end{center}
\end{table*}
\subsection{Performance Metrics} 
Following the standards to evaluate molecular generators, we compare our model with the baselines on the following performance metrics: 
\begin{itemize}
    \item \emph{validity rate}, the ratio of generated molecules that decode to valid SMILES strings, out of the total number of generated molecules;
    \item \emph{novelty rate}, the ratio of valid generated molecules which do not appear in the training set;
    \item \emph{uniqueness rate}, the ratio of unique molecules (not duplicated) out of the total number of valid generated molecules.
\end{itemize}

\begin{figure*}[h]
     \centering
     \begin{subfigure}[b]{0.49\textwidth}
         \centering
         \includegraphics[width=.95\textwidth]{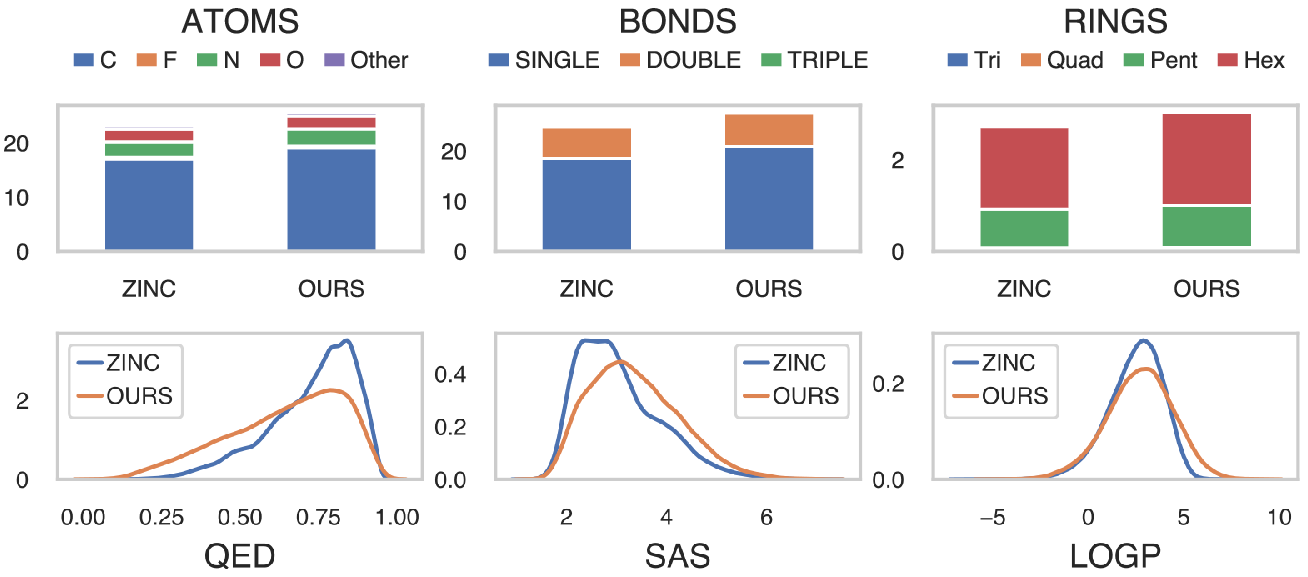}
         \caption{ZINC}
         \label{fig:zinc-props}
     \end{subfigure}
     \hfill
     \begin{subfigure}[b]{0.49\textwidth}
         \centering
         \includegraphics[width=.95\textwidth]{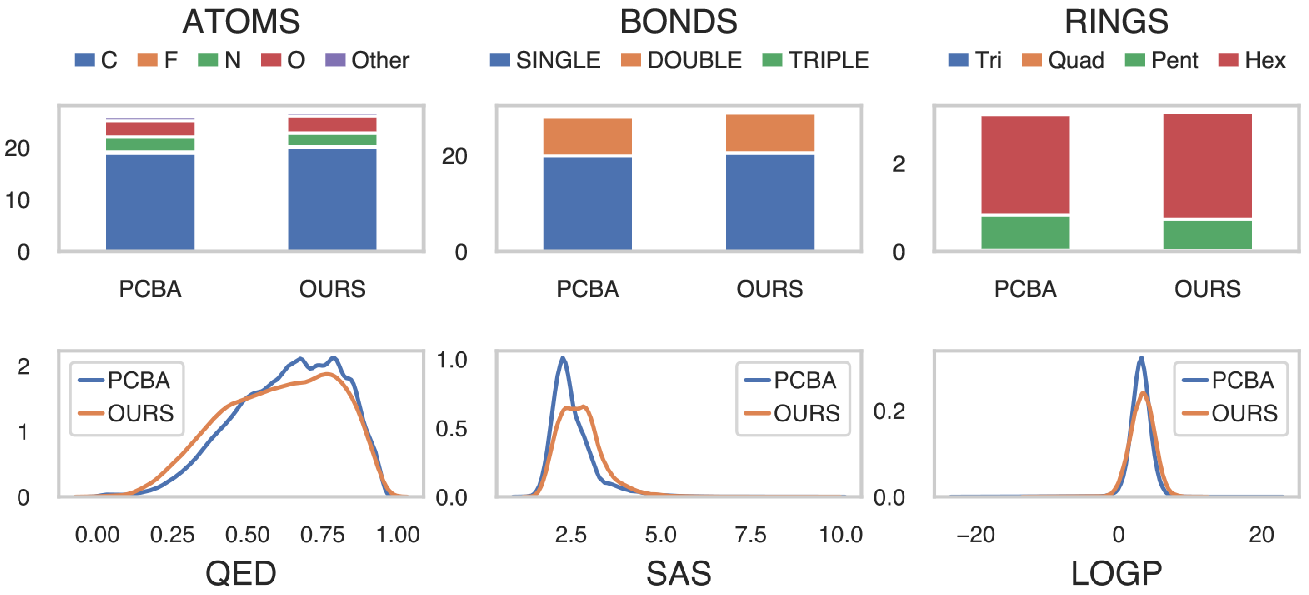}
         \caption{PCBA}
         \label{fig:pcba-props}
     \end{subfigure}
        \caption{Plot of the distributions of structural features (top row) and molecular properties (bottom row) of compounds in the ZINC (a) and PCBA (b) datasets, compared against the 20k compounds sampled from our model.}
        \label{fig:generated-props}
\end{figure*}

\subsection{Baselines}
We compare to baselines found in literature, representing the two families of generative models described in Section~\ref{sec:introduction}. As regards LM-based approaches, we consider ChemVAE \citep{chemvae}, GrammarVAE \citep{grammarvae} and SDVAE \citep{sdvae}, whereas as regards graph-based models, we compare against GraphVAE \citep{graphvae}, CGVAE \citep{cgvae} and NeVAE \citep{nevae}.

\subsection{Hyper-Parameters}
We evaluate our model using the same hyper-parameters for both variants, in order to isolate the effect of our contribution from improvements due to hyper-parameter tuning. We set the embedding dimension to 64, the number of recurrent layers to 2, the number of GRU units per layer to 128 and the latent space size to 100. We used the Adam optimizer with an initial learning rate of 0.00001, annealed every epoch by a multiplicative factor of 0.9, a batch size of 128, and a dropout rate of 0.3 applied to the recurrent layers to prevent overfitting. Training required only 4 epochs: after that, we found empirically that the model started to severely overfit the training set. We used $k=10$ as LFM threshold. The stopping criteria for training is the following: after each epoch, we sample 1000 molecules and measure validity, novelty and uniqueness rates of the sample, stopping whenever the uniqueness rate starts to drop (we found out empirically that samples were stable in terms of validity and novelty rates). After training, we sample 20k molecules for evaluation. We publicly release code and samples for reproducibility\footnote{\scriptsize{\texttt{\url{https://github.com/marcopodda/fragment-based-dgm}}}}. Baseline results are taken from literature\footnote{We found no results in literature for the PCBA dataset.}.

\section{\uppercase{Results}}
The main results of our experiments are summarized in Table~\ref{tab:results}, and provide the answers to the experimental questions posed in Section~\ref{sec:experiments}. As regards Q1, we observe that our model achieves perfect validity scores in the ZINC data, greatly outperforming LM-based models and performing on par with the state of the art. This is true also as regards the PCBA dataset. Since both our variants improve over the LM-based competitors, it is safe to argue that our fragment-based approach can effectively increase validity rates. As regards Q2, we observe an improvement in uniqueness by both our variants, with respect to the LM-based competitors. However, the improvement is noticeably higher whenever the LFM strategy is employed. In the PCBA this trend is even more pronounced. Compared to graph-based models, we see how the model with LFM is now competitive with the state of the art. Lastly, we notice that using LFM yields a small improvement in novelty with respect to the vanilla variant.
\begin{figure*}
     \centering
     \begin{subfigure}[b]{0.49\textwidth}
         \centering
         \includegraphics[width=.8\textwidth]{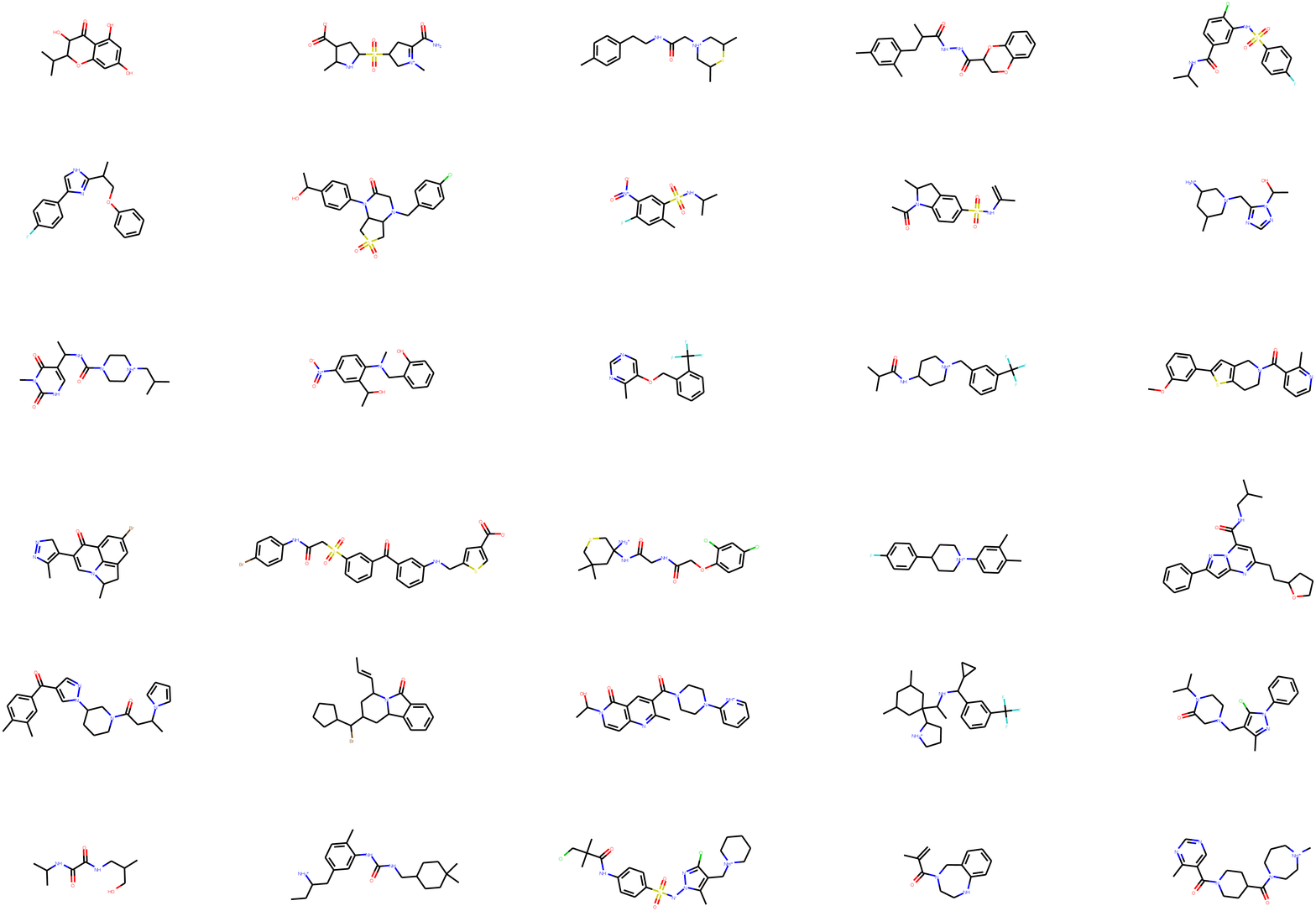}
         \caption{ZINC}
         \label{fig:zinc-samples}
     \end{subfigure}
     \hfill
     \begin{subfigure}[b]{0.49\textwidth}
         \centering
         \includegraphics[width=.8\textwidth]{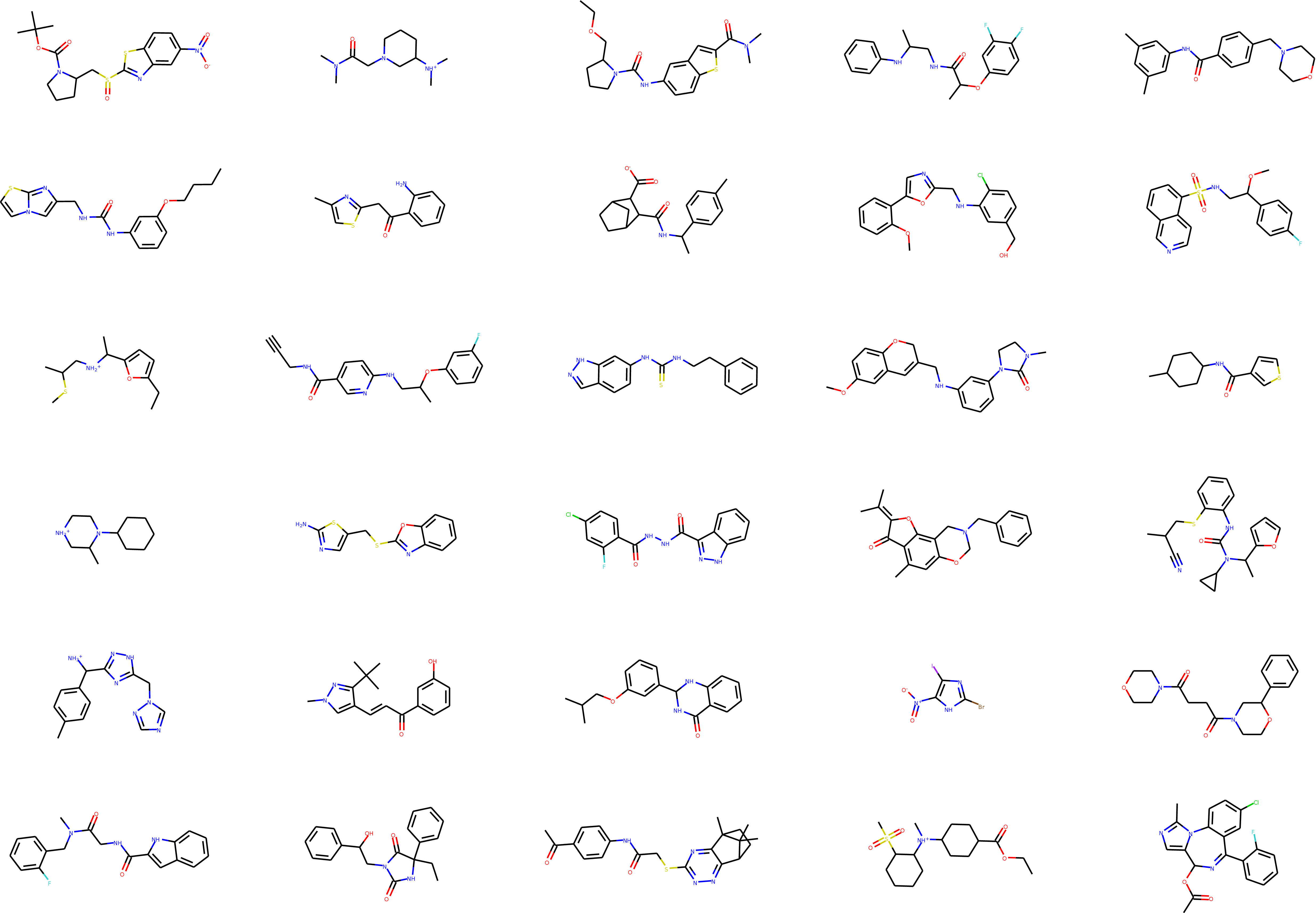}
         \caption{Generated}
         \label{fig:generated-samples}
     \end{subfigure}
        \caption{A random sample of 30 molecules taken from the ZINC dataset (a) and generated by our model (b).}
        \label{fig:samples}
\end{figure*}
\subsection{Molecular Properties of Samples}
One essential aspect of evaluating generative models is determining to what extent generated samples resemble the training data. To this end, we show in Figure~\ref{fig:generated-props} the distribution of several structural features and molecular properties of out-of-dataset samples generated by our model on the ZINC and PCBA datasets, compared to the training sample, after removal of duplicates. Structural features under consideration include atom type counts, bond type counts, and ring type counts (from 3 to 6). As regards molecular properties, we included:
\begin{itemize}
    \item octanol/water Partition coefficient (logP), which measures solubility;
    \item Quantitative Estimate of Drug-likeness \citep{qed} (QED), which measures drug-likeness;
    \item Synthetic Accessibility Score \citep{sas} (SAS), which measures ease of synthesis.
\end{itemize}
In Figure~\ref{fig:generated-props}, our samples against training compounds are compared, as regards the distribution of the three structural features, and molecular properties listed above. Notice that even without the help of an explicit supervision, generated molecules are qualitatively similar to the training data. Figure~\ref{fig:samples} shows two random samples of 30 molecules taken from the ZINC dataset and generated by our model for visual comparison.

\subsection{Computational considerations}
To generate a molecule with $N$ atoms, LM-based methods require $O(C)$ decoding steps, where $C$ is the number of characters in the corresponding SMILES string. Our model requires $O(F)$ decoding steps during generation, where $F$ is the number of fragments (2-3 on average). In contrast, our model requires a substantially larger vocabulary than most LM-based models; its size, however, can be greatly reduced using LFM (e.g. an $\approx 87\%$ reduction for the ZINC dataset). Graph-based methods sample $N$ node embeddings first, then score $O(N^2)$ node pairs to add connections. Moreover, they usually need to enforce chemical validity through additional edge masking. Without masking, performances drop significantly (e.g. NeVAE validity rates drop to 59\%). Our method does not require to enforce validity. 

\subsection{Limitations of the current approach}
We have shown that the presented model is able to perform on par with the state of the art as regards the molecular generation task. At the same time, we acknowledge that it might not be suitable for tasks like molecule optimization in its current form, as the molecular space spanned using LFM is likely less structured than other approaches due to its stochastic component. Given that molecular optimization was outside the scope of this work, we recommend to take this limitation into account when employing our model for that specific task.

\section{\uppercase{Conclusions}}
In this work, we have tackled two main limitations of LM-based generative models for molecules, namely producing chemically invalid as well as duplicate compounds. As regards the first issue, we introduced the first (to our knowledge) fragment-based language model for molecule generation, which operates at fragment level, rather than atom level. As regards the second issue, we presented a low-frequency masking strategy that fosters molecule diversity. In our experiments, we show that our contributions can increase validity and uniqueness rates of LM-based models up to the state of the art, even though an inherently less expressive representation of the molecule is used. As regards future works, we aim at extending this model for task like molecular optimization. This will require the design of novel strategies to maintain high uniqueness rates, while preserving smoothness in latent space. In addition, we would like to adapt the fragment-based paradigm to graph-based molecular generators.

\subsubsection*{Acknowledgements}
This work has been supported by the Italian Ministry of Education, University, and Research (MIUR) under project SIR 2014 LIST-IT (grant n. RBSI14STDE).

\bibliography{references}
\end{document}